\documentclass[sigconf]{acmart}
\AtBeginDocument{%
  }

\copyrightyear{2025}
\acmYear{2025}
\setcopyright{acmlicensed}\acmConference[DFF '25]{Proceedings of the 1st Deepfake Forensics Workshop: Detection, Attribution, Recognition, and Adversarial Challenges in the Era of AI-Generated Media}{October 27--28, 2025}{Dublin, Ireland}
\acmBooktitle{Proceedings of the 1st Deepfake Forensics Workshop: Detection, Attribution, Recognition, and Adversarial Challenges in the Era of AI-Generated Media (DFF '25), October 27--28, 2025, Dublin, Ireland}
\acmDOI{10.1145/3746265.3759663}
\acmISBN{979-8-4007-2047-5/2025/10}

\def\ie{\emph{i.e.}}
\def\eg{\emph{e.g.}}
\def\etal{\emph{et al.}}
 \usepackage{multirow}
\usepackage{graphicx}
\usepackage{subcaption}
\usepackage{xcolor}

\begin{document}

\title{CLIP-Flow: A Universal Discriminator for AI-Generated Images Inspired by Anomaly Detection}

\author{Zhipeng Yuan}
\affiliation{%
  \institution{Jilin University}
  \city{Changchun}
  \state{Jilin}
  \country{China}
}

\author{Kai Wang}
\affiliation{%
  \institution{GIPSA-lab, Univ. Grenoble Alpes, CNRS, Grenoble INP}
  \city{Grenoble}
  \country{France}
}

\author{Weize Quan}
\affiliation{%
  \institution{Institute of Automation, Chinese Academy of Sciences}
  \city{Beijing}
  \country{China}}

\author{Dong-Ming Yan}
\affiliation{%
 \institution{Institute of Automation, Chinese Academy of Sciences}
 \city{Beijing}
 \country{China}}
\email{}

\author{Tieru Wu*}\thanks{*Corresponding author.}
\affiliation{%
  \institution{Jilin University}
  \city{Changchun}
  \state{Jilin}
  \country{China}}
\email{wutr@jlu.edu.cn}
\renewcommand{\shortauthors}{Yuan et al.}

\begin{abstract}
  With the rapid advancement of AI generative models, the visual quality of AI-generated images (AIIs) has become increasingly close to natural images, which inevitably raises security concerns. Most AII detectors often employ the conventional image classification pipeline with natural images and AIIs (generated by a generative model), which can result in limited detection performance for AIIs from unseen generative models. To solve this, we proposed a universal AI-generated image detector from the perspective of anomaly detection. Our discriminator does not need to access any AIIs and learn a generalizable representation with unsupervised learning. Specifically, we use the pre-trained CLIP encoder as the feature extractor and design a normalizing flow-like unsupervised model. Instead of AIIs, proxy images, \eg, obtained by applying a spectral modification operation on natural images, are used for training. Our models are trained by minimizing the likelihood of proxy images, optionally combined with maximizing the likelihood of natural images. Extensive experiments demonstrate the effectiveness of our method on AIIs produced by various image generators. 
\end{abstract}





\maketitle

\section{Introduction}
\label{introduction}

Nowadays, we can easily see fake pictures on the Internet, many of which are synthetic images generated by deep learning algorithms. Indeed, with the rapid development and maturity of deep generative models (such as generative adversarial models and diffusion models), people can easily generate the desired high-quality images. Although image generation technology benefits our lives in some aspects, there will inevitably be people who will use this technology maliciously. For example, AI-generated images (AIIs) can be used for personal and political attacks, making it easier for people to believe the content of fake news. Therefore, it is urgent to develop a universal discriminator of AIIs created by various generative models and tools. 

\begin{figure}[!ht]
    \centering
    \includegraphics[width=0.9\linewidth]{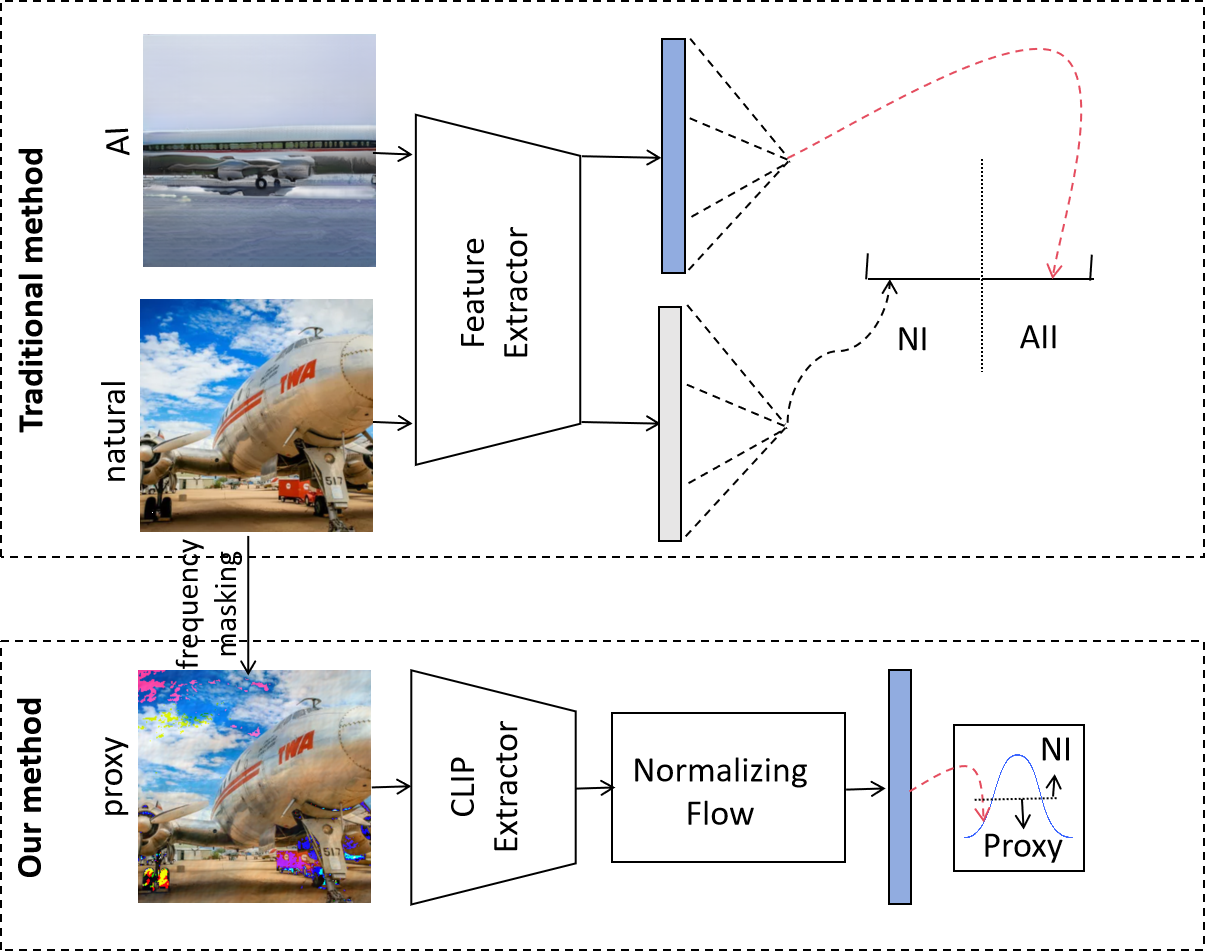}
    \caption{The previous supervised methods used binary classification of natural image (NI) and AI-generated image (AII), and we use the flow model for likelihood optimization of proxy image (optionally also with NI) from the perspective of anomaly detection.}
    \label{fig:difference}
\end{figure}

By choosing a classifier and using two groups of natural images and AIIs as training samples, we can learn an image discriminator in a supervised manner, which is shown in the top of Fig. \ref{fig:difference}.  This approach performs well in identifying generated images similar to those in the training set, but when tested on other types of generated images, the performance can drop drastically. For example, a classifier trained using natural images and ProGAN-generated images \cite{progan} can well identify images generated by ProGAN or generative models with similar structures to ProGAN (such as StyleGAN \cite{stylegan}). However, when faced with images created by other generative models with different structures (such as diffusion models), the classification performance can be very poor. Unfortunately, in the real world, new generative models are constantly emerging, and it is not realistic to update trained detectors at all times to deal with these new models. Our goal in this paper is to achieve high generalization performance for the AII detector, \ie, with high classification accuracy on AIIs created by unknown generative models, by exploring a new approach inspired by anomaly detection.

As mentioned above and observed in the influential paper of \cite{ojha2023fakedetect}, the supervised learning approach may have limitations in learning only the very specific artifacts present in the AIIs in the training set, leading to limited generalization capability. This has motivated us to tackle the detector generalization problem by exploiting a different learning mechanism of anomaly detection \cite{pang2021deep}. In existing anomaly detection methods, training is usually performed only on normal samples and any sample that deviates significantly from the normal samples will be classified as an anomaly. Intuitively, in our AII detection task, normal samples correspond to natural images. Given this idea, we started by considering using only natural images to train anomaly detection methods. Unfortunately, limited generalization performance has been achieved with various anomaly detection algorithms and feature extractors, including the well-known CLIP embedding extractor \cite{radford2021learning} which has proven its great utility in AII detection as shown in \cite{ojha2023fakedetect}. One possible explanation for this limited generalization is the difficulty in using existing anomaly detection algorithms to capture the highly diverse representations of natural images. This has led us to reconsider the problem formulation and to better adapt the anomaly detection mechanism to our task of AII detection.

We assume that natural images follow an underlying distribution that is different from AIIs. As mentioned previously, it is difficult to identify AIIs by modeling this distribution, which has motivated us to rethink the problem from a new perspective with a focus on the so-called proxy images whose distribution deviates from that of natural images. Indeed, it is believed that despite the high visual quality, synthetic images in general exhibit deviations in the spatial domain (\eg, different correlations between pixel values) \cite{quan_2024_cgformer} or the frequency domain (\eg, artifacts in certain frequency bands) \cite{cai2021frequency,doloriel2024frequency}. Following this idea, we apply spatial or spectral operation (\eg, the random frequency masking \cite{doloriel2024frequency} which shows the best performance in our study) on natural images to construct a group of processed images, regard them as anomalous samples, extract their CLIP features, and train a detector by minimizing their likelihood based on a normalizing flow-like model, a popular tool for anomaly detection \cite{kobyzev2020normalizing,yu2021fastflow,cflow} (bottom of Fig. \ref{fig:difference}). Well-constructed proxy images are expected to be closer to AIIs than to natural images in the feature space. This ``unconventional'' anomaly detection framework called \textit{CILP-Flow} shows very good performance on benchmark test sets of natural image and AII classification, notably by not using any AII during training. We further improve the generalization performance by proposing a self-supervised (in a loose sense) framework in which proxy images and their corresponding natural images are used together to train the flow model (still without using any AII during training), with minimization and maximization of the likelihood of the two types of images, respectively.

Our contributions are summarized as follows: 
\vspace{-0.7mm}
\begin{itemize}
    \item To the best of our knowledge, we conducted a careful and exploratory study on the usefulness of methods in the field of anomaly detection to solve the problem of AII detection.
    \item When we only use anomalous frequency-masked proxy images for training, we achieve very competitive AII detection performance on popular benchmark test sets with an mAP of 93.85\% and an accuracy of 84.93\%.
    \item Our model has good flexibility and can also be trained in a self-supervised manner: when trained using both natural images and proxy images, the model achieves higher detection performance with an mAP of 95.56\% and an accuracy of 87.30\%.
    \item We proposed a novel approach to solving the AII detection problem that is inspired by anomaly detection, and we hope that our work can inspire future efforts in this promising research direction.
\end{itemize}


\section{Related Work}
\label{relatedwork}

\subsection{AI-Generated Image Detection}


\textbf{Spatial-Based Approaches.} This kind of methods applies deep models to learn discrimination features in the pixel space directly. Wang \etal~\cite{cnn-detect} explored two simple image augmentation operations (\ie, JPEG compression and Gaussian blur) to improve the CNN-based detector. Chai \etal~\cite{patchforensics} constructed a patch-based AII detector using the truncated part of CNN backbones (\eg, Xception and ResNet50). Natraj \etal~\cite{natarajan} combined the co-occurrence matrix of the input pixels and CNN to distinguish between natural images and AIIs, while Bai \etal~\cite{bai2021robust} identified AI-generated images based on co-occurrence relations in the feature channels of CNN model. Ojha \etal~\cite{ojha2023fakedetect} applied the ViT encoder of the pre-trained CLIP to detect fake images and obtained decent generalization performance. With the help of a pre-trained diffusion model, Wang \etal~\cite{wang_2023_dire} extracted the reconstruction error between the input image and its reconstructed version to distinguish between real images and diffusion-generated images.

\textbf{Frequency-Based Approaches.} This type of methods identifies fake images based on revealing traces in the frequency domain. Zhang \etal~\cite{zhang2019gan} and Frank \etal~\cite{frank_freq} designed classifiers that take the frequency spectrum as input, attempting to capture the artifacts caused by the up-sampler of GAN models. Durall \etal~\cite{Durall2020watch} and Dzanic \etal~\cite{Dzanic2020Fourier} developed AI-generated image discriminators based on the difference between natural images and AIIs in the high-frequency component. Liu \etal~\cite{liu_2022_detect} utilized the frequency spectrum information of learnable image noise patterns to detect AIIs. Recently, authors of \cite{doloriel2024frequency} and \cite{sythesisimage} explored frequency-based modification to improve the generalization of an AII detector with traditional binary cross-entropy loss.

\subsection{Anomaly Detection}

Anomaly detection usually requires that only normal samples are used during training, and the model is asked to distinguish whether a sample is normal or anomalous during testing. Typical applications of anomaly detection include industrial defect detection, disease detection, video surveillance, \textit{etc.} \cite{pang2021deep}. Recent anomaly detection methods for images can be roughly grouped into three main categories: reconstruction-based methods, distillation-based methods, and density-based methods.

\textbf{Reconstruction-Based Methods.} This kind of methods reconstructs normal samples in the training stage, and classifies samples that cannot be well reconstructed in testing as anomalies. GANomaly \cite{akcay2019ganomaly} uses the GAN structure as a reconstruction model. DRAEM \cite{zavrtanik2021draem} uses an autoencoder as a reconstruction model and, for better performance, also trains a discriminator to identify artificially generated anomalous samples using Perlin noise.

\textbf{Distillation-Based Methods.} STFPM \cite{wang2103student} attempts to distinguish anomalous samples by training a student model to learn the intermediate layer features extracted by a teacher model for normal samples. After such a distillation procedure, the student model tends to have a higher loss on anomalous samples. RD \cite{deng2022anomaly} is another distillation-based method, where a one-class embedding is learned to represent normal data.

\textbf{Density-Based Methods.} Such methods extract features of normal samples and model the distribution of these features. An anomaly score can be derived usually based on the probability of a sample under the learned distribution. Different methods mainly differ in terms of feature extractor and distribution modeling \cite{ahuja2019probabilistic,defard2021padim}. For example, PaDiM \cite{defard2021padim} detects anomalies by extracting features of local image patches and then learning a multivariate Gaussian distribution of normal patches during training. FastFlow \cite{yu2021fastflow} and CFlow \cite{cflow} use the normalizing flow, a powerful generative model based on neural network, for the distribution learning. Considering the conceptional and computational simplicity of density-based approaches compared to other methods, we use the normalizing flow model in our work for AI-generated image discrimination.

\section{Proposed Method}
\label{proposedmethod}

\begin{figure}
    \centering
    \includegraphics[width=1.0\linewidth]{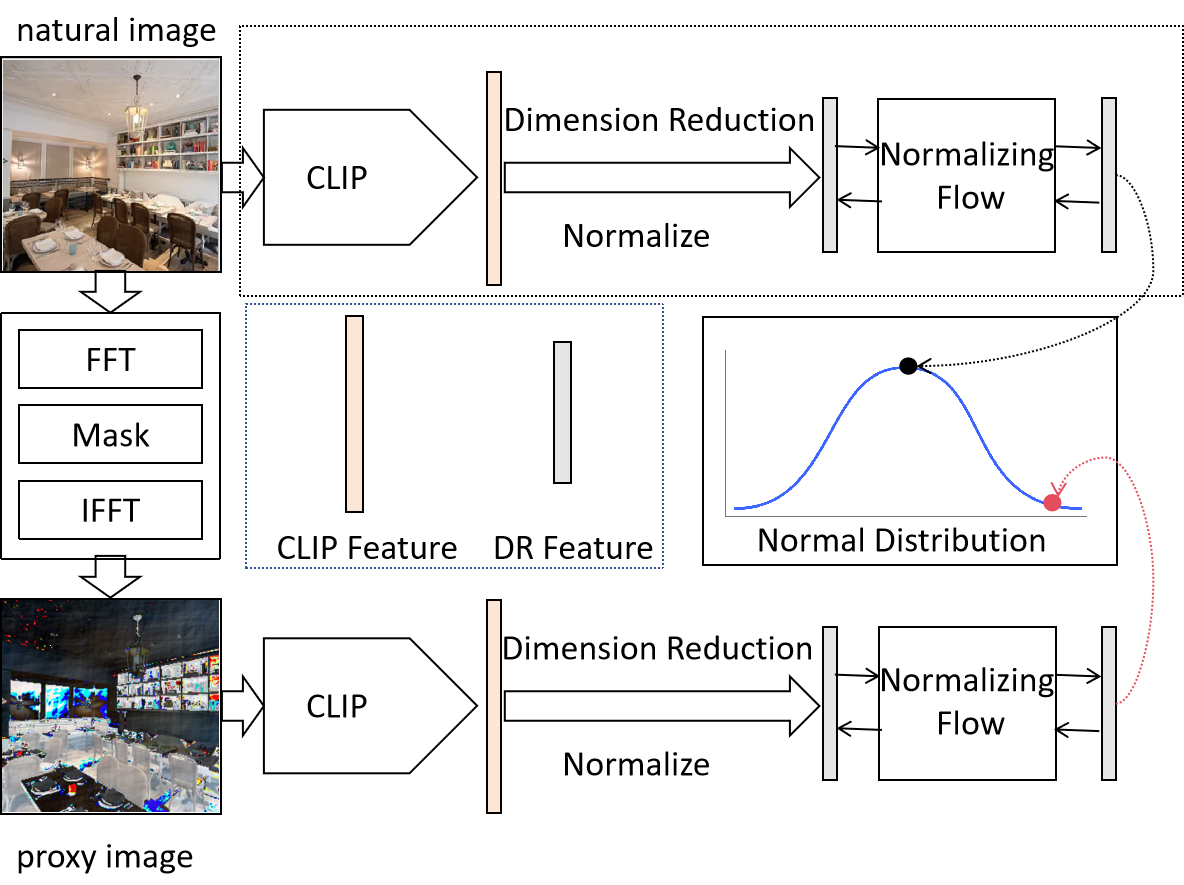}
    \caption{Framework of CLIP-Flow.}
    \label{fig:Framework}
\end{figure}

The proposed method is illustrated in Fig. \ref{fig:Framework}. Unlike previous supervised detectors, our method does not need any AI-generated image during training. Instead, we create and use proxy images, \eg, obtained by applying random frequency masking on natural images. We then use the CLIP model to extract features from the proxy images, adapt them with dimension reduction (DR) and normalization, and turn the distribution of adapted features into a normal distribution using unsupervised normalizing flow. Our model can also be jointly trained with unprocessed natural images (upper part of Fig. \ref{fig:Framework}). During testing, we can calculate an anomaly score based on the flow model to classify images.

\subsection{Feature Extractor}
\label{clipextractor}

In many computer vision tasks, the ViT encoder of the large-scale pre-trained CLIP model \cite{radford2021learning} is often used as the feature extraction backbone. Similarly, Ojha \etal~\cite{ojha2023fakedetect} and Li and Wang \cite{sythesisimage} observed superior forensic performance in AI-generated image detection with CLIP features compared to other popular image embeddings. We have similar observations in our preliminary experiments when combining different image features with the flow model. Therefore, in order to construct a universal AII discriminator, in our work, we also choose the ViT encoder of CLIP to extract the image feature. Particularly, ViT extracts a high-dimensional feature from an image $x$, and we reduce its dimension to balance the performance and the computational cost. We then normalize it, and the feature adaptation process is written as follows: 
\begin{equation}
  z= \frac{DR(\phi (x))}{\| DR(\phi(x)) \|_2},
  \label{eq:feature extractor}
\end{equation}
where $\phi(.)$ is the CLIP feature extraction function and $DR(.)$ is the dimension reduction operation with a linear layer.

\subsection{Proxy Image Construction}
\label{proxyimage}

We have attempted different methods to construct proxy images. Comparative experimental results are presented in Section \ref{experiments}, which show that operations in either spatial or spectral domain can work well in our proposed framework and that the frequency masking operation described below achieves the best detection performance.

There are many works that exploit frequency information for AII detection, with the basic idea that there is a telltale difference between natural and generated images in the frequency domain of the image. We also use frequency information, more specifically by applying random frequency masking \cite{doloriel2024frequency} to process the original image. This is combined with a novel learning framework inspired by anomaly detection to achieve good generalization and flexibility. Let $x$ be a natural image, we first use the Fourier transform to convert it to the frequency domain. Then, we produce a random binary mask and use it to mask a ratio of a certain part (low, mid, high) of the frequency spectrum (please refer to \cite{doloriel2024frequency} for more details). We convert this masked frequency spectrum back to RGB space with the inverse Fourier transform to obtain a proxy image $x^{'}$. The whole process is as follows:
\begin{equation}
  x^{'}=ifft(mask \odot fft(x)),
  \label{eq:frequency masking}
\end{equation}
where $fft(.)$ represents Fast Fourier transform, $mask$ is a random binary mask, $\odot$ is the element-wise multiplication, and $ifft(.)$ represents inverse Fast Fourier transform.

As mentioned in Section \ref{introduction}, performance remains limited if we train only on natural images $x$ with anomaly detection algorithms, possibly due to the high diversity of natural images and the blindness regarding the anomalies to be detected. Instead, decent performance is achieved if we train a flow model only with proxy images $x^{'}$ by minimizing their likelihood values (bottom branch of Fig. \ref{fig:Framework}). This implies that frequency-masked proxy images are in general closer to AIIs than to natural images in the feature space. Performance can be further improved if we use both natural images $x$ and proxy images $x^{'}$ for training (top and bottom branches of Fig. \ref{fig:Framework}), by maximizing and minimizing the likelihood value, respectively. This can be loosely considered as a self-supervised learning method that demonstrates strong zero-shot performance on the AII detection task. Both learning methods are inspired by anomaly detection and, to the best of our knowledge, are new in the AII detection literature.

\begin{table*}[t]
\caption{The performance (in AP, \%) of our model and existing methods. In the last group of rows, N means model trained only on natural images with likelihood maximization, P means model trained only on proxy images with likelihood minimization, and N+P means model trained on both types of images above with likelihood maximization and minimization respectively. The symbols are the same in the following tables.}
  {\small
    \centering
    \tabcolsep=0.1cm
    \resizebox{1.\linewidth}{!}{
    \begin{tabular}{cc cccccc c cc cc c ccc ccc c c}
    \toprule

        \multirow{2}{*}{\shortstack[c]{Methods}} & \multirow{2}{*}{Variant} & \multicolumn{6}{c}{Generative Adversarial Networks} & \multirow{2}{*}{\shortstack[c]{Deep\\Fakes}} & \multicolumn{2}{c}{Low level vision} & \multicolumn{2}{c}{Perceptual loss} &\multirow{2}{*}{Guided} & \multicolumn{3}{c}{LDM} & \multicolumn{3}{c}{Glide} & \multirow{2}{*}{DALL-E} & Total \\
    \cmidrule(lr){3-8} \cmidrule(lr){10-11} \cmidrule(lr){12-13} \cmidrule(lr){15-17} \cmidrule(lr){18-20} \cmidrule(lr){22-22}

    & & \shortstack[c]{Pro-\\GAN} & \shortstack[c]{Cycle-\\GAN} & \shortstack[c]{Big-\\GAN} & \shortstack[c]{Style-\\GAN} & \shortstack[c]{Gau-\\GAN} &  \shortstack[c]{Star-\\GAN}   &  & SITD & SAN & CRN & IMLE & & \shortstack[c]{200\\steps} & \shortstack[c]{200\\w/ CFG} & \shortstack[c]{100\\steps} & \shortstack[c]{100\\27} & \shortstack[c]{50\\27} & \shortstack[c]{100\\10} & & \shortstack[c]{mAP}
    
    \\ 

    \midrule

\multirow{11}{*}{\shortstack[c]{Anomaly\\Detection}} & PaDiM & 43.04 & 40.86& 47.06 & 45.58 & 39.86 & 42.13 & 46.21 & 48.40 & 54.34 & 42.18 & 39.22 & 41.04 & 52.62 & 59.57 & 51.56 & 40.77 & 38.81 & 41.03 & 73.61 & 46.73 \\
& CFA & 57.62 & 66.60 & 56.41 & 71.13 & 68.63 & 70.23 & 41.54 & 66.38 & 51.65 & 77.34 & 83.02 & 68.64 & 54.50 & 46.34 & 53.94 & 69.44 & 74.36 & 69.00& 40.95&62.51  \\
& GANomaly &  48.93 & 49.39 & 52.33 & 44.37 & 45.80 & 49.51 & 52.97 & 49.74 & 62.29 & 37.57 & 38.88 & 39.49 & 50.41 & 56.47 & 48.60 & 39.71 & 37.01 & 40.67 & 56.38 & 47.39\\
& DSR &  50.35 & 45.80 & 50.04 & 48.46 & 51.29 & 45.79 & 56.47 & 47.79 & 52.07 & 40.44 & 54.58 & 47.48 & 45.86 & 48.02 & 45.60 & 46.87 & 45.37 & 49.70 & 47.36 & 48.39 \\
& DRAEM &  92.73 & 43.24 & 59.50 & 94.27 & 51.48 & 97.15 & 48.54 & 41.53 & 51.09 & 47.00 & 40.52 & 41.73 & 95.28 & 90.72 & 95.33 & 43.59 & 43.54 & 42.67 & 98.35 & 64.12 \\
& STFPM &  53.60 & 55.16 & 47.89 & 45.88 & 53.32 & 62.27 & 51.11 & 52.85 & 58.08 & 44.64 & 49.31 & 41.79 & 51.69 & 53.11 & 51.11 & 43.77 & 41.05 & 46.75 & 60.58 & 50.73 \\
& RD & 47.46 & 46.54 & 48.43 & 40.60 & 41.96 & 49.92 & 49.52 & 42.62 & 54.34 & 34.29 & 32.21 & 38.32 & 47.02 & 43.06 & 47.28 & 34.74 & 33.45 & 33.85 & 50.11 & 42.93 \\
& DFM & 49.00 & 50.92 & 47.02 & 48.86 & 46.67 & 50.00 & 49.69 & 49.16 & 53.43 & 45.71 & 42.57 & 47.90 & 59.60 & 58.85 & 59.79 & 52.64 & 49.52 & 52.18 & 59.25 & 51.20 \\
& FastFlow & 49.08 & 49.49 & 44.68 & 45.81 & 47.58 & 49.37 & 53.00 & 48.30 & 55.53 & 62.88 & 61.13 & 48.58 & 54.58 & 54.95 & 55.12 & 54.95 & 53.69 & 55.01 & 64.40 & 53.06 \\
& CFlow & 48.97 & 55.10 & 51.05 & 50.44 & 47.34 & 45.87 & 48.09 & 84.65 & 49.38 & 45.88 & 53.92 & 40.88 & 75.21 & 59.88 & 76.73 & 50.40 & 47.72 & 44.60 & 56.43 & 54.33 \\

\midrule

\multirow{3}{*}{\shortstack[c]{CNNSpot}} & Blur+JPEG (0.1) & \textbf{100.00} & 93.47 & 84.50 & 99.54 & 89.49 & 98.15 & 89.02 & 73.75 & 59.47 & 98.24 & 98.40 & 73.72 & 70.62 & 71.00 & 70.54 & 80.65 & 84.91 & 82.07 & 70.59 & 83.58 \\
& Blur+JPEG (0.5) & \textbf{100.00} & 96.83 & 88.24 & 98.29 & 98.09 & 95.44 & 66.27 & 86.00 & 61.20 & \textbf{98.94} & \textbf{99.52} & 68.57 & 66.00 & 66.68 & 65.39 & 73.29 & 78.02 & 76.23 & 65.93 & 81.52\\
& ViT:CLIP (B+J 0.5) &  99.98 & 93.32 & 83.63 & 88.14 & 92.81 & 84.62 & 67.23 & \textbf{93.48} & 55.21 & 88.75 & 96.22 & 55.74 & 52.52 & 54.51 & 52.20 & 56.64 & 61.13 & 56.64 & 62.74 & 73.44 \\ 

\midrule

\multirow{2}{*}{\shortstack[c]{PatchForensics}} & ResNet50-Layer1  &  98.86 & 72.04 & 68.79 & 92.96 & 55.90 & 92.06 & 60.18 & 65.82 & 52.87 & 68.74 & 67.59 & 70.05 & 87.84 & 84.94 & 88.10 & 74.54 & 76.28 & 75.84 & 77.07 & 75.28\\
& Xception-Block2 &  80.88 & 72.84 & 71.66 & 85.75 & 65.99 & 69.25 & 76.55 & 76.19 & 76.34 & 74.52 & 68.52 & 75.03 & 87.10 & 86.72 & 86.40 & 85.37 & 83.73 & 78.38 & 75.67 & 77.73\\

\midrule

\multirow{1}{*}{\shortstack[c]{Co-Occur}} & -  &  99.74 & 80.95 & 50.61 & 98.63 & 53.11 & 67.99 & 59.14& 68.98 & 60.42 & 73.06 & 87.21 & 70.20 & 91.21 & 89.02 & 92.39 & 89.32 & 88.35 & 82.79 & 80.96 & 78.11\\

\midrule

\shortstack[c]{FreqSpec} & CycleGAN & 55.39 & \textbf{100.00} & 75.08 & 55.11 & 66.08 & \textbf{100.00} & 45.18 & 47.46 & 57.12 & 53.61 & 50.98 & 57.72 & 77.72 & 77.25 & 76.47 & 68.58 & 64.58 & 61.92 & 67.77 & 66.21\\

\midrule

\multirow{1}{*}{\shortstack[c]{DIRE}} & - & \textbf{100.00} & 77.12 & 76.30 & 92.64 & 71.08 & \textbf{100.00} & \textbf{95.32} & 62.02 & 63.30 & 95.77 & 94.03 & 83.34 & 95.19 & 93.89 & 95.12 & 93.78 & 95.56 & 95.99 & 61.27 & 86.41 \\
\midrule

\multirow{1}{*}{\shortstack[c]{UnivFD}} & LC & \textbf{100.00} & 99.46 & \textbf{99.59} & 97.24 & \textbf{99.98} & 99.60 & 82.45 & 61.32 & 79.02 & 96.72 & 99.00 & 87.77 & 99.14 & 92.15 & 99.17 & 94.74 & 95.34 & 94.57 & 97.15 & 93.38\\
\midrule

\multirow{1}{*}{\shortstack[c]{Freq-mask}} & CLIP & \textbf{100.00} & 99.04 & 96.89 & 92.66 & 99.83 & 98.86 & 77.17 & 67.41 & 75.80 & 78.83 & 95.85 & 81.77 & 95.45 & 80.93 & 96.48 & 90.66 & 91.71 & 90.23 & 86.73 & 89.28\\
\midrule

\multirow{1}{*}{\shortstack[c]{OnlyReal}} & - & 99.99 & \textbf{100.00} & 99.48 & \textbf{99.77}& 99.96 & \textbf{100.00} & 94.75 & 40.65 & 68.01 & 89.25 & 98.96 & \textbf{89.93} & \textbf{99.90} & \textbf{99.64} & \textbf{99.91} & 96.55 & 97.60 & 96.36 & \textbf{99.99} & 93.18\\
\midrule

\multirow{3}{*}{\textbf{\shortstack[c]{Ours}}} & N & 87.99 & 63.86 & 74.00 & 57.69 & 63.39 & 93.40 & 66.89 &40.37 & 51.34 & 37.26 & 45.58 & 61.18 & 42.24 & 42.90 & 42.50 & 34.73 & 35.75 & 34.79 & 54.43 & 54.23\\
& P & 99.86 & 98.40 & 98.38 & 89.42 & 99.84 & 98.13 & 92.23 & 84.18 & 72.09 & 98.14 & 98.79 & 76.29 & 98.49 & 94.33 & 98.46 & 95.49 & 97.07 & 96.48 & 97.15 & 93.85\\
& N+P &  99.94 & 99.87 & 99.24 & 97.10 & 99.92 & 99.50 & 87.34 & 80.06 & \textbf{79.48} & 94.85 & 96.94 & 87.78 & 99.41 & 96.93 & 99.44 & \textbf{99.35} & \textbf{99.59} & \textbf{99.28} & 99.55 & \textbf{95.56} \\ 

    \bottomrule
    
    \end{tabular}}
    }
\label{tab:ap}
\end{table*}

\subsection{Normalizing Flow and Training Objective}

For distribution learning and sample likelihood estimation, we choose the normalizing flow model \cite{dinh2017density,papamakarios2021normalizing}, which is a popular probabilistic method for generative modeling. It is able to map a simple probability distribution (such as the Gaussian distribution) to a complex distribution via a series of reversible transformations with the trick of change of variable. With the flow model, it is also possible to transform a complex distribution into a simple one, \eg, the Gaussian distribution. In our CLIP-Flow method, we assume that the distribution of the adapted CLIP feature $z$ of natural image follows a (complex) distribution as $z \sim p_Z(z)$, the chosen simple distribution is denoted $u \sim p_U(u)$, and the reversible bijective transformation between the two distributions is denoted $g^{-1}: Z \rightarrow U$. Then, the log-likelihood of any $z \in Z$ can be estimated as: 
\begin{equation}
\begin{aligned}
  \log p_{Z}(z,\theta) & = \log p_{U}(u) + \log |\det(J)|\\
                       & = \log p_{U}(g^{-1}(z)) + \log |\det(J)|,
\end{aligned}
\label{eq:likelihood}
\end{equation}
where the matrix $J=\nabla_z g^{-1}(z;\theta)$ is the Jacobian of a bijective invertible flow model parameterized by $\theta$.

We then calculate for $z$ a KL divergence to measure the difference between its distribution and the approximated distribution with the transformation of flow model: 
\begin{equation}
\begin{aligned}    
D_{KL}[p_{Z}^{*}(z)||p_{Z}(z)]&= - \mathbb{E}_{p_{Z}^{*}(z)}[\log p_{Z}(z;\theta)-\log p_{Z}^{*}(z)]   \\
 &=- \mathbb{E}_{p_{Z}^{*}(z)}[\log p_{Z}(z;\theta)] + c_1,
\end{aligned}
\label{eq:KL1}
\end{equation}
where $p_{Z}^{*}$ represents the true distribution of $z$, $p_{Z}$ represents the approximated distribution by using the flow model parameterized by $\theta$, and $c_1$ is a term independent of $\theta$. 

Here we propose a \textit{new} flow model which can handle two types of features having ``complementary'' distributions. More specifically, for feature $z^{'}$ of proxy image, we assume that it follows a distribution related to that of natural image as $z^{'} \sim K/p_Z(z^{'})$ (under the mild condition of $p_Z(z^{'}) \neq 0$, $\forall z^{'}$), where $K$ is a normalization constant to ensure a valid probability density function integrating to 1. This is a reasonable assumption because at certain position in feature space with high probability density of natural image, it certainly leads to low probability density for artificial proxy image at that position, and \textit{vice versa}. Similar to Eq. (\ref{eq:KL1}), we have: 
\begin{equation}
\begin{aligned}
D_{KL}[p_{Z^{'}}^{*}(z^{'})||\frac{K}{p_{Z}(z^{'})}] & = -\mathbb{E}_{p_{Z^{'}}^{*}(z^{'})}[\log(\frac{K}{p_{Z}(z^{'};\theta)})] + c_2 \\
 & =\mathbb{E}_{p_{Z^{'}}^{*}(z^{'})}[\log p_{Z}(z^{'};\theta)] + c_3,
\end{aligned}
\label{eq:KL2}
\end{equation}
where $p_{Z^{'}}^{*}$ represents the true distribution of $z^{'}$, $K/p_{Z}$ is the approximated distribution by using the flow model parameterized by $\theta$, and $c_2$ and $c_3$ are terms independent of $\theta$.

For the flow model to work properly, we need to minimize the KL divergences in Eqs. (\ref{eq:KL1}) and (\ref{eq:KL2}), which elegantly leads to a maximization of the log-likelihood of natural images and a minimization of the log-likelihood of proxy images, according to these two equations (please note their different signs for the non-constant term). Therefore, we use the likelihood estimation to obtain the following loss function to minimize: 
\begin{equation}
\begin{aligned}
    \mathcal{L}(\theta) = -\frac{1}{N_{o}}\sum^{N_{o}}_{n=1}\log p_{Z}(z_{n};\theta)+\frac{1}{N_{p}}\sum^{N_{p}}_{m=1} \log p_{Z}(z^{'}_{m};\theta),
\end{aligned}
\label{Loss}
\end{equation}
where $N_{o}$ is the number of natural images, $N_{p}$ is the number of proxy images, and $z_{n}$ and $z^{'}_{m}$ represent respectively the feature of the natural and proxy image. The first term on the right side of Eq. (\ref{Loss}) means to maximize the likelihood of natural images, while the second term means to minimize the likelihood of proxy images. We have a so-called self-supervised training when we use both terms. In contrast, when we use only one type of images for unsupervised training, one of the two terms vanishes, \eg, the first term vanishes when we train the flow model using only proxy images.

We set the transformed distribution as a multidimensional Gaussian distribution $u \sim \mathcal{N}(0,I)$. At the same time, since the loss is the sum of each dimension, following \cite{cflow} we divide it by the number of dimensions $C$ to normalize the loss. Finally, the training objective of the flow model is to minimize the following loss function: 
\begin{equation}
\begin{aligned}
    Loss = &-\frac{1}{2N_{o}}\sum^{N_{o}}_{n=1} \frac{\|u_n\|_2^2 - 2\log |\det(J)|}{C} \\
    &+ \frac{1}{2N_{p}}\sum^{N_{p}}_{m=1} \frac{\|u^{'}_m\|_2^2 -2 \log|\det (J)|}{C}, 
\end{aligned}
    \label{finalloss}
\end{equation}
where $u_n=g^{-1}(z_n;\theta)$ and $u^{'}_m=g^{-1}(z^{'}_m;\theta)$ are the transformed features of the flow model corresponding to the natural image and the proxy image, respectively.

\subsection{Anomaly Score}

Our goal is to determine whether an image is a natural image or an AII. For this purpose and as shown in Eq. (\ref{anomalyscore}) below, we naturally define and use an anomaly score based on the \textit{negative} log-likelihood of the sample. The normalized log-likelihood term in Eq. (\ref{anomalyscore}) is exactly the same as that used in the flow model loss function in Eq. (\ref{finalloss}), where we minimize (or maximize) the log-likelihood of proxy images (or natural images). Therefore, natural images tend to have low values of this score (because of the negative sign before the log-likelihood term in the definition of the anomaly score), while AI-generated images tend to have high values. 
\begin{equation}
score = - \frac{\log p_{Z}(z;\theta)}{C} = \frac{\|u\|_2^2 -2\log |\det (J)|}{2C}.
\label{anomalyscore}
\end{equation}

\section{Experiments}
\label{experiments}


\subsection{Experimental Settings}
\label{experiments_settings}

\textbf{Datasets.} Following the representative methods on AII detection~\cite{cnn-detect,ojha2023fakedetect}, we used natural images from LSUN for training, with a size of $256 \times 256$ pixels and 20 categories. Detection models are evaluated on the benchmark (in total 19 datasets) shared by the above two references~\cite{cnn-detect,ojha2023fakedetect}, including six GAN models: ProGAN, CycleGAN, BigGAN, StyleGAN, GauGAN, and StarGAN; DeepFakes; two low-level vision models: SITD and SAN; two perceptual loss-based models: CRN and IMLE; three diffusion models: Guided, LDM (latent diffusion model) with three different generation modes, and Glide with three different variants; DALL-E. These tested samples have diverse content and can be used to comprehensively assess the performance of detection models.

\textbf{Metrics.} Following~\cite{cnn-detect,ojha2023fakedetect}, we use the average precision (AP) and classification accuracy for comparison. AP is a comprehensive reflection of the classification ability of our anomaly scores. In order to find a fixed threshold to calculate the accuracy, we created a validation set. Positive samples are LSUN natural images that are neither in the training set nor in the testing set, and negative samples are frequency-perturbed images by masking the frequency band of (30, 100), generated in a different way than the frequency masking used for the training set.

\textbf{Implementation Details.} Our model is implemented with PyTorch and trained on NVIDIA GeForce RTX GPU. We use Adam optimizer with a learning rate of $10^{-4}$ and batch size of 128. CLIP: ViT-L/14 is the feature extractor. The bijection mapping in the flow model is implemented with an affine coupling layer~\cite{dinh2017density}. Proxy images are generated by random frequency masking \cite{doloriel2024frequency} with the masking ratio of 1.0 (for P mode, our method trained only with proxy images) and 0.1 (for N+P mode, our method trained with both natural and proxy images) in the low part of the spectrum. Experimentally, for P mode, it is advantageous to use proxy images with large changes, in order to confidently minimize likelihood values on them. For N+P mode, since it is a self-supervised approach, we can use proxy images with small changes for strict distinction under the setting of a hard discrimination problem. Proxy images constructed with frequency masking achieve the best AII detection performance, and the results obtained with other types of proxy images are presented and compared later in this section. During training, all images are resized to $256 \times 256$ with bilinear interpolation and then randomly cropped to $224 \times 224$. The normalization uses the same mean and variance as in the CLIP model \cite{radford2021learning}. The P mode takes 30 epochs and the N+P mode takes 10 epochs. During testing, the central $224 \times 224$ crop of the $256 \times 256$ resized image is used for evaluation.

\begin{table*}[t]
\caption{The performance (in Accuracy, \%) of our model and existing methods.}
  {\small
    \centering
    \tabcolsep=0.1cm
    \resizebox{1.\linewidth}{!}{
    \begin{tabular}{cc cccccc c cc cc c ccc ccc c c}
    \toprule

        \multirow{2}{*}{\shortstack[c]{Methods}} & \multirow{2}{*}{Variant} & \multicolumn{6}{c}{Generative Adversarial Networks} &\multirow{2}{*}{\shortstack[c]{Deep\\Fakes}} & \multicolumn{2}{c}{Low level vision} & \multicolumn{2}{c}{Perceptual loss} &\multirow{2}{*}{Guided} & \multicolumn{3}{c}{LDM} & \multicolumn{3}{c}{Glide} & \multirow{2}{*}{DALL-E} & Total \\
    \cmidrule(lr){3-8} \cmidrule(lr){10-11} \cmidrule(lr){12-13} \cmidrule(lr){15-17} \cmidrule(lr){18-20} \cmidrule(lr){22-22}

    & & \shortstack[c]{Pro-\\GAN} & \shortstack[c]{Cycle-\\GAN} & \shortstack[c]{Big-\\GAN} & \shortstack[c]{Style-\\GAN} & \shortstack[c]{Gau-\\GAN} &  \shortstack[c]{Star-\\GAN}   &  & SITD & SAN & CRN & IMLE & & \shortstack[c]{200\\steps} & \shortstack[c]{200\\w/ CFG} & \shortstack[c]{100\\steps} & \shortstack[c]{100\\27} & \shortstack[c]{50\\27} & \shortstack[c]{100\\10} & & \shortstack[c]{Avg.\\acc}
    
    \\ 

    \midrule

\multirow{4}{*}{\shortstack[c]{CNNSpot}} & Blur+JPEG (0.1) & 99.99 & 85.20 & 70.20 & 85.70 & 78.95 & 91.70 & 53.47 & 66.67 & 48.69 & 86.31 & 86.26 & 60.07 & 54.03 & 54.96 & 54.14 & 60.78 & 63.80 & 65.66 & 55.58 & 69.58 \\
& Blur+JPEG (0.5) & \textbf{100.00} & 80.77 & 58.98 & 69.24 & 79.25 & 80.94 & 51.06 & 56.94 & 47.73 & 87.58 & 94.07 & 51.90 & 51.33 & 51.93 & 51.28 & 54.43 & 55.97 & 54.36 & 52.26 & 64.73 \\
& Oracle (B+J 0.5) & \textbf{100.00} & 90.88 & 82.40 & 93.11 & 93.52 & 87.27 & 62.48 & \textbf{76.67} & 57.04 & \textbf{95.28} & \textbf{96.93} & 65.20 & 63.15 & 62.39 & 61.50 & 65.36 & 69.52 & 66.18 & 60.10 & 76.26 \\ 
& ViT:CLIP (B+J 0.5) & 98.94 & 78.80 & 60.62 & 60.56 & 66.82 & 62.31 & 52.28 & 65.28 & 47.97 & 64.09 & 79.54 & 50.66 & 50.74 & 51.04 & 50.76 & 52.15 & 53.07 & 52.06 & 53.18 & 60.57 \\ 

\midrule

\multirow{2}{*}{\shortstack[c]{PatchForensics}} & ResNet50-Layer1  &  94.38 & 67.38 & 64.62 & 82.26 & 57.19 & 80.29 & 55.32 & 64.59 & 51.24 & 54.29 & 55.11 & 65.14 & 79.09 & 76.17 & 79.36 & 67.06 & 68.55 & 68.04 & 69.44 & 68.39\\
& Xception-Block2 &  75.03 & 68.97 & 68.47 & 79.16 & 64.23 & 63.94 & 75.54 & 75.14 & \textbf{75.28} & 72.33 & 55.30 & 67.41 & 76.50 & 76.10 & 75.77 & 74.81 & 73.28 & 68.52 & 67.91 & 71.24\\



\midrule

\multirow{1}{*}{\shortstack[c]{Co-Occur}} & -  & 97.70 & 63.15 & 53.75 & 92.50 & 51.10 & 54.70 & 57.10 & 63.06 & 55.85 & 65.65 & 65.80 & 60.50 & 70.70 & 70.55 & 71.00 & 70.25 & 69.60 & 69.90 & 67.55 & 66.86\\

\midrule

\shortstack[c]{FreqSpec} & CycleGAN & 49.90 & \textbf{99.90} & 50.50 & 49.90 & 50.30 & 99.70 & 50.10 & 50.00 & 48.00 & 50.60 & 50.10 & 50.90 & 50.40 & 50.40 & 50.30 & 51.70 & 51.40 & 50.40 & 50.00 & 55.45\\

\midrule

{\shortstack[c]{DIRE}} & - & \textbf{100.00} & 67.50 & 64.53 & 83.40 & 63.03 & \textbf{99.95} & 93.27 & 56.67 & 63.47 & 55.54 & 55.55 & 75.35 & 88.15 & 89.10 & 89.50 & 87.15 & 90.35 & 91.40 & 62.90 & 77.73 \\

\midrule

\multirow{1}{*}{\shortstack[c]{UnivFD}} & LC & \textbf{100.00} & 98.50 & \textbf{94.50} & 82.00 & \textbf{99.50} & 97.00 & 66.60 & 63.00 & 57.50 & 59.50 & 72.00 & 70.03 & 94.19 & 73.76 & 94.36 & 79.07 & 79.85 & 78.14 & 86.78 & 81.38 \\

\midrule

\multirow{1}{*}{\shortstack[c]{OnlyReal}} & - & 99.15 & 93.50 & 89.90 & \textbf{98.05} & 80.50 & 99.60 & \textbf{85.00} & 49.50 & 71.50 & 50.10 & 50.10 & \textbf{81.10} & \textbf{97.95} & \textbf{97.40} & \textbf{98.10} & 88.40 & 91.90 & 87.20 & \textbf{98.10} & 84.58 \\

\midrule

\multirow{3}{*}{\textbf{\shortstack[c]{Ours}}} & N & 81.75 & 54.85 & 58.84 & 59.46 & 56.38 & 50.06 & 49.94 & 50.00 & 52.05 & 50.00 & 50.00 & 53.65 & 49.05 & 48.00 & 48.45 & 46.97 & 47.30 & 46.30 & 51.60 & 52.88 \\
& P &  95.76 & 91.74 & 92.72 & 75.09 & 98.26 & 90.26 & 83.16 & 76.11 & 56.62 & 92.61 & 93.34 & 60.55 & 92.45 & 80.40 & 92.80 & 82.94 & 86.75 & 84.75 & 87.30 & 84.93\\
&N+P& 99.22 & 93.67 & 90.80 & 84.42 & 88.11 & 96.65 & 66.77 & 54.44 & 68.49 & 84.79 & 85.12 & 78.55 & 96.80 & 88.35 & 96.50 & \textbf{96.30} & \textbf{97.25} & \textbf{96.10} & 96.55 & \textbf{87.30}\\

    \bottomrule
    
    \end{tabular}}
    }
\label{tab:main_comparison}
\end{table*}

\subsection{Comparison with Existing Methods}

In this subsection, we compare our method with several advanced anomaly detection approaches: PaDiM~\cite{defard2021padim}, CFA~\cite{lee2022cfa}, GANomaly~\cite{akcay2019ganomaly},
DSR~\cite{zavrtanik2022dsr},
DRAEM~\cite{zavrtanik2021draem}, STFPM~\cite{wang2103student}, RD~\cite{deng2022anomaly},
DFM~\cite{ahuja2019probabilistic},
FastFlow~\cite{yu2021fastflow}, CFlow~\cite{cflow}, and existing AI-generated image detectors: CNNSpot~\cite{cnn-detect}, PatchForensics~\cite{patchforensics}, Co-Occur~\cite{natarajan}, FreqSpec~\cite{zhang2019gan}, DIRE~\cite{wang_2023_dire}, UnivFD~\cite{ojha2023fakedetect}, Freq-mask \cite{doloriel2024frequency} and OnlyReal~\cite{sythesisimage}. The numerical results are reported in Table~\ref{tab:ap}. For anomaly detection methods, we train them completely according to the settings and instructions of the original models using only natural images in our training set and test them on the AII detection benchmark test sets. For most AI-generated image detection methods, we use the results as presented in the UnivFD paper \cite{ojha2023fakedetect}. Besides, we also compare with DIRE \cite{wang_2023_dire}, Freq-mask \cite{doloriel2024frequency} and OnlyReal \cite{sythesisimage} (all based on supervised learning algorithm), with the same setting of UnivFD.

From the first part of Table \ref{tab:ap}, we can see that existing anomaly detection methods have limited performance on the task of AI-generated image detection, with mAP remaining quite low around 50\% for the vast majority of tested methods (last column of Table \ref{tab:ap}). These results are the outcome of our extensive study on the performance evaluation of recent anomaly detection methods on AII detection, to our knowledge the first time in the literature for such a study. Possible explanations of the limited performance, as mentioned before, are the insufficient ability of existing methods to capture the very diverse content of natural images and the blindness to anomalies to be detected. The DRAEM method \cite{zavrtanik2021draem} shows slightly better performance than other anomaly detection methods with mAP of around 64\%. Upon inspection, we hypothesize that this better performance is mainly due to the construction of some artificial anomaly samples in the DRAEM method. This is in part in line with the idea of constructing proxy images in our method, which nevertheless shows much better AII detection performance, as presented in the following.

In the last group of rows of Table \ref{tab:ap}, we show the test results of three modes of our model. N represents our model trained with likelihood maximization on 7,200 natural images, P represents our model trained with likelihood minimization on 7,200 proxy images, and N+P represents our model trained with both the likelihood maximization on natural images and likelihood minimization on proxy images (in total 14,400 images). We can see that when we only maximize the likelihood of natural images, the performance is not ideal, again probably due to the reasons mentioned in the last paragraph for existing anomaly detection methods. On the contrary, when we only minimize the likelihood of proxy images, the performance of our model has been significantly improved. The mAP of our method has reached a high value of 93.85\%, while not using any AI-generated image for training. This implies that, despite their simplicity of construction, proxy images have distinctive differences from natural images, making them closer in the feature space to AIIs generated by a variety of generative models, including GANs and diffusion models, than to natural images. Finally, if we jointly maximize the likelihood of natural images with the N+P mode, the mAP of our model can be further improved to 95.56\%. Towards the end of Table \ref{tab:ap}, we compare with the very recent method OnlyReal \cite{sythesisimage}, which, similar to our method, does not use AIIs to build a forensic detector. However, it is worth noting that the essential structure of OnlyReal remains a supervised binary model with the conventional cross-entropy loss and consistent with traditional AII detection methods, and that it cannot be trained using only a single class of images. In contrast, our method can be effectively trained in an unsupervised manner using only a single class of proxy images, making it more flexible and more inspiring for future research. Regarding the AII detection performance, we can see from Table \ref{tab:ap} that OnlyReal achieves best performance on many test sets (such as CycleGAN, StarGAN and LDM), while the performance of our method (N+P mode) is very close on these datasets. On datasets where OnlyReal performs poorly (such as SITD and CRN), our model still performs well, so the final average performance mAP of our method is higher than OnlyReal (95.56\% vs. 93.18\%).

\begin{figure}[t]
    \centering
    \begin{subfigure}[b]{0.50\textwidth}
    \includegraphics[width=1.0\linewidth]{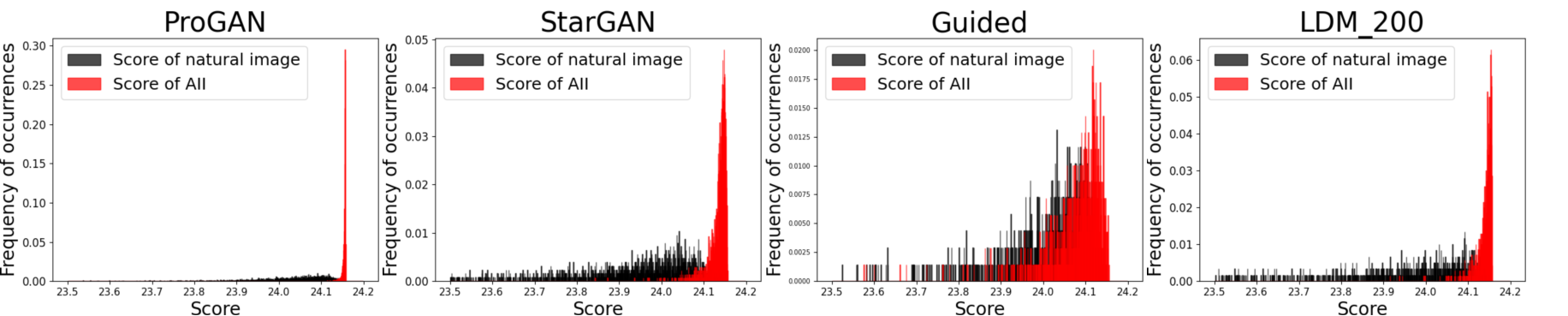}
    \end{subfigure}
\hfill
    \begin{subfigure}[b]{0.50\textwidth}
        \centering
    \includegraphics[width=1.0\linewidth]{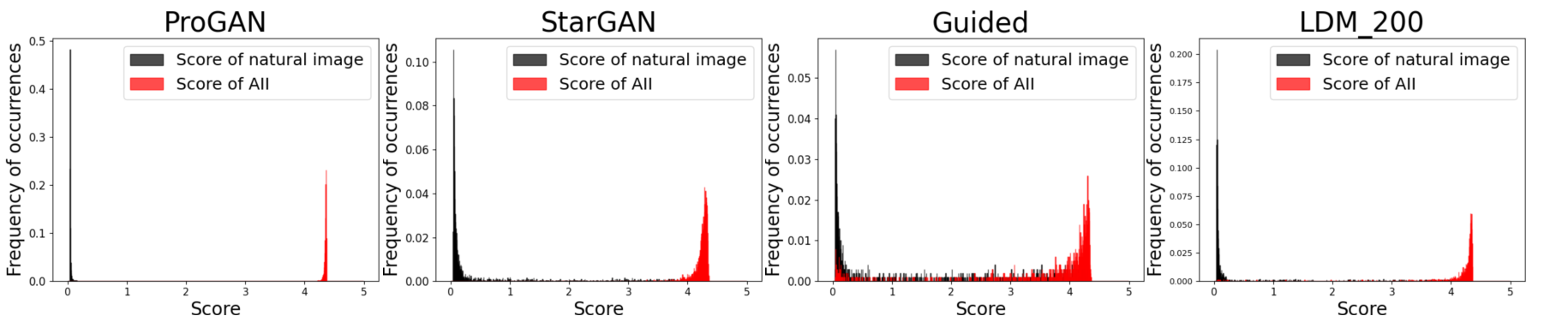}
    \end{subfigure}
\caption{Histograms of anomaly scores on test sets of ProGAN, StarGAN, Guided, and LDM\_200 (from left to right), for the P mode (top row) and the N+P mode (bottom row) of our method. The range of score values is different for the two modes due to different training losses (\textit{cf}. main text). The scores of the proxy images, not shown here for the sake of brevity, are all densely clustered at very high values.}
\label{fig:histogram}
\end{figure}

In Fig. \ref{fig:histogram}, we show histograms of the anomaly scores of some test sets. It can be seen that, for ProGAN, StarGAN and LDM\_200, with our method trained with P mode, the histograms of natural images and AIIs remain different, but they are close and have small overlap; when we use our model trained with N+P mode, the two histograms are very well separated with a gap between them. Similar to existing methods, our model has limited performance on Guided; nevertheless, the histogram separability is much improved with N+P mode than with P mode (third column of Fig. \ref{fig:histogram}). The magnitude of the anomaly scores in the two modes is different, as shown in Fig. \ref{fig:histogram}, and this is mainly due to the different training losses used in the two modes. For the P mode, the training strategy is to minimize the log-likelihood (thus maximizing the anomaly score) of the proxy images, while for the N+P mode, we additionally maximize the log-likelihood of the natural images. The more complex training loss in the N+P mode experimentally leads to anomaly scores of lower magnitude, probably due to the need to reach a trade-off between the losses of the two types of images.

Table \ref{tab:main_comparison} presents the test accuracy results of our model and representative existing methods. For our model, we choose a threshold based on a synthetic validation set as mentioned in Section \ref{experiments_settings}. The average accuracy reaches 84.93\% for N mode of our model, which is 3.55\% higher than the reference UnivFD method.  When we maximize the likelihood of natural images and minimize the likelihood of proxy images at the same time, \ie, N+P mode, the average accuracy of our model improves further and reaches 87.30\%.

\begin{table*}[ht!]
\caption{The performance (in AP, \%) of different types of proxy images. FM stands for frequency masking.}
  {\small
    \centering
    \tabcolsep=0.1cm
    \resizebox{1.\linewidth}{!}{
    \begin{tabular}{cc cccccc c cc cc c ccc ccc c c}
    \toprule

        \multirow{2}{*}{\shortstack[c]{Mode}} & \multirow{2}{*}{\shortstack[c]{Type}} & \multicolumn{6}{c}{Generative Adversarial Networks} &\multirow{2}{*}{\shortstack[c]{Deep\\Fakes}} & \multicolumn{2}{c}{Low level vision} & \multicolumn{2}{c}{Perceptual loss} &\multirow{2}{*}{Guided} & \multicolumn{3}{c}{LDM} & \multicolumn{3}{c}{Glide} & \multirow{2}{*}{DALL-E} & Total \\
    \cmidrule(lr){3-8} \cmidrule(lr){10-11} \cmidrule(lr){12-13} \cmidrule(lr){15-17} \cmidrule(lr){18-20} \cmidrule(lr){22-22}

    & & \shortstack[c]{Pro-\\GAN} & \shortstack[c]{Cycle-\\GAN} & \shortstack[c]{Big-\\GAN} & \shortstack[c]{Style-\\GAN} & \shortstack[c]{Gau-\\GAN} &  \shortstack[c]{Star-\\GAN}   &  & SITD & SAN & CRN & IMLE & & \shortstack[c]{200\\steps} & \shortstack[c]{200\\w/ CFG} & \shortstack[c]{100\\steps} & \shortstack[c]{100\\27} & \shortstack[c]{50\\27} & \shortstack[c]{100\\10} & & \shortstack[c]{mAP}
    
    \\ 

    \midrule

\multirow{6}{*}{\shortstack[c]{N+P}} & Smoothing & 96.41 & 68.84 & 74.09 & 84.56 & 88.84 & 48.36 & 42.69 & 67.86 & 59.21 & 80.09 & 85.49 & 61.61 & 89.19 & 78.36 & 88.48 & 83.72 & 85.67 & 84.02 & 83.13 & 76.25 \\ 

& Sharpening &  99.66 & 96.22 & 98.59 & 88.49 & 97.94 & \textbf{99.74} & 79.05 & 77.89 & 67.66 & 44.84 & 80.11 & 79.33 & 96.99 & 92.87 & 96.63 & 65.93 & 70.97 & 69.34 & 99.56 & 84.31\\

& Gaussian noise & 99.73 & 96.35 & 96.82 & 88.82 & 98.88 & 98.89 & 69.30 & 73.18 & 63.22 & 90.33 & 96.83 & 73.54 & 95.93 & 90.42 & 95.80 & 70.16 & 78.13 & 73.36 & 99.13 & 86.78 \\

& Color jitter &  99.87 & 98.73 & 97.90 & \textbf{98.15} & 97.88 & 99.57 & 81.71 & 70.94 & 71.05 & 76.70 & 96.79 & 82.18 & \textbf{99.50} & \textbf{98.10} & \textbf{99.45} & 97.79 & 98.52 & 98.17 & \textbf{99.88} & 92.78\\

& FM (onlyphase) & \textbf{99.96} & 99.86 & \textbf{99.31} & 96.11 & \textbf{99.93} & 99.73 & 85.78 & 79.52 & 79.31 & \textbf{95.39} & \textbf{97.62} & 87.66 & 99.20 & 95.98 & 99.14 & 98.93 & 99.32 & 98.87 & 99.32 & 95.31\\

& FM &  99.94 & \textbf{99.87} & 99.24 & 97.10 & 99.92 & 99.50 & \textbf{87.34} & \textbf{80.06} & \textbf{79.48} & 94.85 & 96.94 & \textbf{87.78} & 99.41 & 96.93 & 99.44 & \textbf{99.35} & \textbf{99.59} & \textbf{99.28} & 99.55 & \textbf{95.56} \\ 

    \bottomrule

    \end{tabular}}
    }
\label{tab:aug}
\end{table*}

\begin{table*}[ht!]
\caption{The performance (in AP, \%) of proxy image generation with the masking of different frequency parts. We randomly set the low, mid, and high part of the image spectrum (as defined in \protect\cite{doloriel2024frequency}) to 0 with a probability of 0.1.}
  {\small
    \centering
    \tabcolsep=0.1cm
    \resizebox{1.\linewidth}{!}{
    \begin{tabular}{cc cccccc c cc cc c ccc ccc c c}
    \toprule

        \multirow{2}{*}{\shortstack[c]{Mode}} & \multirow{2}{*}{\shortstack[c]{Frequency\\Part}} & \multicolumn{6}{c}{Generative Adversarial Networks} &\multirow{2}{*}{\shortstack[c]{Deep\\Fakes}} & \multicolumn{2}{c}{Low level vision} & \multicolumn{2}{c}{Perceptual loss} &\multirow{2}{*}{Guided} & \multicolumn{3}{c}{LDM} & \multicolumn{3}{c}{Glide} & \multirow{2}{*}{DALL-E} & Total \\
    \cmidrule(lr){3-8} \cmidrule(lr){10-11} \cmidrule(lr){12-13} \cmidrule(lr){15-17} \cmidrule(lr){18-20} \cmidrule(lr){22-22}

    & & \shortstack[c]{Pro-\\GAN} & \shortstack[c]{Cycle-\\GAN} & \shortstack[c]{Big-\\GAN} & \shortstack[c]{Style-\\GAN} & \shortstack[c]{Gau-\\GAN} &  \shortstack[c]{Star-\\GAN}   &  & SITD & SAN & CRN & IMLE & & \shortstack[c]{200\\steps} & \shortstack[c]{200\\w/ CFG} & \shortstack[c]{100\\steps} & \shortstack[c]{100\\27} & \shortstack[c]{50\\27} & \shortstack[c]{100\\10} & & \shortstack[c]{mAP}
    
    \\ 

    \midrule

\multirow{3}{*}{\shortstack[c]{N+P}} & Low  &  99.94 & \textbf{99.87} & \textbf{99.24} & 97.10 & \textbf{99.92} & \textbf{99.50} & 87.34 & 80.06 & \textbf{79.48} & \textbf{94.85} & 96.94 & \textbf{87.78} & \textbf{99.41} & 96.93 & 99.44 & \textbf{99.35} & \textbf{99.59} & \textbf{99.28} & 99.55 & \textbf{95.56} \\ 
& Mid &  99.50 & 98.34 & 97.38 & \textbf{98.83} & 98.51 & 96.31 & 73.16 & 74.52 & 57.31 & 90.51 & 95.70 & 75.96 & 99.39 & \textbf{98.07} & \textbf{99.45} & 97.98 & 98.66 & 98.25 & \textbf{99.79} & 91.98\\
&High& \textbf{99.96} & 99.81 & 99.12 & 97.46 & 99.88 & 99.30 & \textbf{88.80} & \textbf{80.43} & 78.31 & 94.64 & \textbf{97.01} & 86.36 & 99.40 & 97.04 & 99.41 & 99.20 & 99.51 & 99.12 & 99.54 & 95.49 \\

    \bottomrule

    \end{tabular}}
    }
\label{tab:fre_aug}
\end{table*}

\subsection{Additional Results and Ablation Studies}

\textbf{Other Types of Proxy Images.} As mentioned in Section \ref{proxyimage}, we have also attempted to use proxy images constructed by other types of operations, including spatial operations such as smoothing, sharpening, Gaussian noise addition, and color jitter, as well as a variant of frequency masking that randomly masks only the phase while keeping the spectral amplitude unchanged. For the sake of brevity, we show in Table \ref{tab:aug} the results in the N+P mode of our detector. The parameter setting of the new operations is similar to that of frequency masking, \ie, for the N+P mode a relatively small modification is introduced with appropriate parameters. For example, Gaussian noise was added with a moderate standard deviation of 5 (pixel value between 0 and 255) to produce noisy proxy images. It can be seen from Table \ref{tab:aug} that routine spatial operations (smoothing, sharpening, and noise addition) result in limited detection performance, while color jitter leads to much better overall mAP. One possible explanation is that color jitter introduces a plausible color deviation from natural images. We notice that only masking the spectrum phase achieves a high mAP, very close to the frequency masking of both the phase and the amplitude. This observation is interesting and remains understandable because the spectrum phase is proven to carry important structural and perceptual information \cite{oppenheim1981phase,ni2007phase,gladilin2015phase}. It is also encouraging to see that our method can cope well with both spatially (color jitter) and spectrally (frequency masking) modified proxy images for the training of an effective AII detector, while not mimicking specific artifacts of AIIs. Future research shall be devoted to understanding the effect of different operations and to constructing more effective proxy images.

\textbf{Spectrum Part for Frequency Masking.} In our work, we generate proxy images by randomly masking the low part of the image spectrum. What effect would masking information on the high part and the mid part have? The corresponding results are reported in Table \ref{tab:fre_aug}. We can see that our method can perform properly in all three cases, with the low and high parts having a higher mAP than the mid part. AIIs of some models such as ProGAN, GauGAN, IMLE, and Glide can be distinguished very well when we mask either of the three parts of the spectrum to generate proxy images. AIIs of some other models such as DeepFakes, SAN, and Guided are much better distinguished when we mask the low or high part than the mid part. In general, the detection performance on each generative model and with each frequency part is different, and masking the low and high parts has better overall performance than masking the mid part, with the low part masking being slightly better than the high part masking. Therefore, we choose to perform random frequency masking in the low part of the spectrum of natural images to produce proxy images.

\begin{figure}
    \centering
    \includegraphics[width=0.99\linewidth]{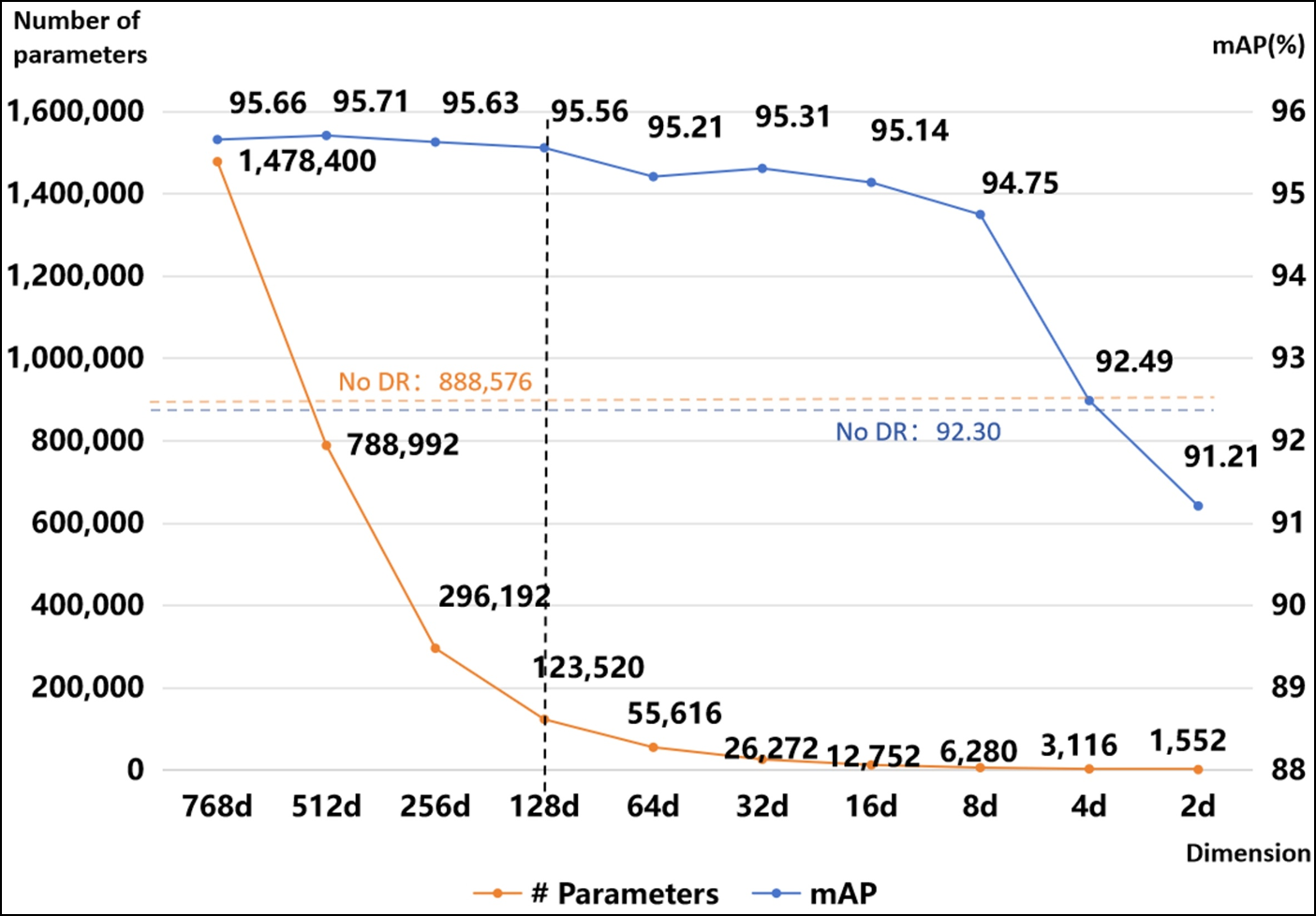}
    \caption{Results of ablation studies regarding the DR layer.}
    \label{fig:DR dimensions}
\end{figure}

\textbf{Dimension Reduction.} We conducted ablation experiments to verify the impact of dimension reduction (DR) of the CLIP feature. In the experiments, we simultaneously maximized the likelihood of natural images and minimized the likelihood of proxy images, and kept other parameters unchanged while only varying the output dimension of the DR layer. As shown in the results of Fig. \ref{fig:DR dimensions}, if the DR layer is not used (the two ``No DR'' horizontal lines), not only the number of parameters will be very large, but also the model performance will decrease. Using the DR layer to reduce the feature dimension simplifies the model and improves the performance. Regarding the choice of the reduced feature dimension, we chose 128 which has a good trade-off between model complexity (its number of parameters) and performance (its mAP).

\textbf{Feature Normalization.} After the CLIP feature dimension is reduced by the linear layer of DR, we also normalize its magnitude to 1. Our intuitions behind this normalization are as follows: 1) The CLIP model was trained by contrastive learning with a cosine similarity loss, so its features are in theory independent of their magnitude; 2) With feature normalization, the input distribution of the flow model is constrained on a unit hyper-sphere, which would be less complicated to learn than an arbitrary distribution and thus reduce the risk of overfitting. Experimentally, the mAP is 93.41\% without normalization, while the mAP reaches 95.56\% (an increase of 1.15\%) with feature normalization.

\subsection{Evaluation on Emerging Generative Models}

To further examine generalization, we built new test datasets based on prompt-driven image generation. Specifically, from real images, we used a vision-to-language model to generate prompts, then synthesized corresponding fake images using state-of-the-art generators, including DALL·E 3, DeepFloyd, RealVisXL v4.0, and Stable Diffusion v1.5. Each new dataset comprises 1000 real and 1000 generated images. We compare our method with UnivFD~\cite{ojha2023fakedetect}. As shown in Table~\ref{tab:custom_results}, our model (P mode, without any fine-tuning, \textit{i.e.}, the same model evaluated in previous subsections) achieves high AP (Average Precision) scores across the new test datasets. These emerging and unseen generators produce high-quality, semantically rich images that pose substantial challenges for AII detectors. Overall, our method demonstrates strong detection performance even on unseen and recently released generative models, suggesting enhanced generalization to real-world deployment settings. At last, we would like to clarify that our goal in this work is to explore AII detection from a new anomaly detection perspective. We do not aim to outperform every concurrent and recent method (mostly based on supervised learning), and we do not claim SOTA performance.

\begin{table}[t!]
\centering
\caption{AP (\%) on datasets from recent generative models. Our method has strong generalization to unseen generators.}
\label{tab:custom_results}
\small
\begin{tabular}{l|cc}
\toprule
\textbf{Generator} & \textbf{Ours} & \textbf{UnivFD} \\
\midrule
DALL{\small\textperiodcentered}E 3 & 90.48 & 61.78 \\
DeepFloyd        & 97.17 & 86.47 \\
RealVisXL v4.0      & 90.70 & 61.98 \\
Stable Diffusion v1.5 & 94.93 & 85.96 \\
\midrule
\textbf{Average}    & \textbf{93.32} & \textbf{74.05} \\
\bottomrule
\end{tabular}
\end{table}

\section{Conclusion}
\label{conclusion}

Unlike the conventional AII detection pipeline that relies on supervised learning, we propose a universal and generalizable framework inspired by anomaly detection. Our model includes a CLIP-based feature extractor and a normalization flow-like module. Without requiring access to AI-generated images, our model is trained using anomalous frequency-masked images and optionally natural images by optimizing the image likelihood under the flow model. Experimental results demonstrate the strong generalization capability of our CLIP-Flow model on various AII detection benchmark datasets. In particular, our work shows the feasibility of leveraging anomaly detection with unsupervised or self-supervised learning for universal detection of AI-generated images. An interesting by-product of our study is a new flow model that can handle two types of samples of complementary distributions, and this new model could be useful in other data analysis applications. In the future, we would like to explore other proxy image generation approaches, propose new anomaly detection methods that can cope better with the AII detection problem, and extend our framework to other multimedia forensic challenges. Code is available at https://github.com/Yzp1018/CLIP-Flow.

\section*{Acknowledgments}

This work is partially funded by the French National Research Agency (Grants numbers ANR-23-IAS4-0004-02 and ANR-15-IDEX-02) and the Excellent Youth Program of State Key Laboratory of Multimodal Artificial Intelligence Systems.

\bibliographystyle{ACM-Reference-Format}
\bibliography{refs_final}










\end{document}